\documentclass{article}

\usepackage[sglblindworkshop, final]{ai4nextg_neurips_2025}

\usepackage[utf8]{inputenc}   
\usepackage[T1]{fontenc}      

\usepackage[english]{babel}

\usepackage{amsmath,amssymb,amsthm,bm,amsfonts}

\usepackage{graphicx}                   
\usepackage[table,xcdraw]{xcolor}       
\usepackage[font=footnotesize]{caption}
\usepackage{subcaption}
\usepackage{booktabs,multirow,tabularx} 
\usepackage{tikz}
\usetikzlibrary{patterns,arrows.meta}
\usepackage{wrapfig}

\usepackage[detect-all]{siunitx}        
\usepackage{pifont}                     
\usepackage{enumitem}                   
\usepackage{xspace}                     
\usepackage{nicefrac}                   
\usepackage[bottom]{footmisc}           
\usepackage{balance}                    
\usepackage{verbatim}                   
\usepackage{listings}               
\usepackage{tikz}
\usetikzlibrary{arrows.meta,positioning,fit,shapes.multipart}

\usepackage{microtype}                  
\usepackage{url}                        

\usepackage{hyperref}

\newcommand{\eg}{{\sl e.g., }}

\newcommand{\BfPara}[1]{{\noindent\bf#1.}\xspace}

\newcounter{descriptcount}
\newlist{enumdescript}{description}{1}
\setlist[enumdescript,1]{%
  before={\setcounter{descriptcount}{0}%
  \renewcommand*\thedescriptcount{\arabic{descriptcount}}},%
  font={\bfseries\stepcounter{descriptcount}Q\thedescriptcount:}%
}

\definecolor{color1}{RGB}{228,26,28}
\definecolor{color2}{RGB}{55,126,184}
\definecolor{color3}{RGB}{77,175,74}
\definecolor{color4}{RGB}{152,78,163}
\definecolor{color5}{RGB}{255,127,0}
\definecolor{color6}{RGB}{200,200,200}

\title{PEARL: Peer-Enhanced Adaptive Radio via On-Device LLM}

\author{%
  \href{https://juhyung-lee.com/}{Ju-Hyung Lee}\thanks{Equal contribution. \{juhyung.lee, yanqing.lu\}@nokia.com} \,$^{1}$ \quad
  \href{https://yqlu1015.github.io/}{Yanqing Lu}\footnotemark[1] \,$^{1,2}$ \quad
  \href{https://scholar.google.com/citations?user=A9FYSF8AAAAJ}{Klaus Doppler}$^{1}$ \\
  $^{1}$Nokia Technologies, Sunnyvale, CA, USA \\
  $^{2}$Department of Computer Science, University of Southern California, Los Angeles, CA, USA
}

\begin{document}

\maketitle

\begin{abstract}
We present \textbf{PEARL} (\textit{Peer-Enhanced Adaptive Radio via On-Device LLM}), a framework for cooperative cross-layer optimization in device-to-device (D2D) communication. 
Building on our previous work on single-device on-device LLMs \citep{lee2025ondevicellm}, PEARL extends the paradigm by leveraging both publisher and subscriber states to guide \texttt{Wi-Fi Aware} (WA) parameter selection. 
A context-aware reward, which normalizes latency by application tolerances and modulates energy by device battery states, provides richer supervision for KL-based fine-tuning. 
We study two lightweight variants: PEARL (Head + Low-Rank Adaptation (LoRA) \citep{hu2022lora}) achieves the best overall performance, while PEARL-Lite (Head-only) delivers sub-20 ms inference at near-identical objective scores.
Across synthetic scenarios grounded in real measurements, PEARL improves objective scores over heuristic and compact model baselines and reduces energy by up to 16\% in cooperative low-battery cases. 
These results demonstrate that peer-aware context, reward-aligned training, and head-based efficiency make LLMs practical for always-on, on-device cross-layer control. 
Code, real-world demo, and dataset are available at \url{https://github.com/abman23/pearl}.
\end{abstract}

\section{Introduction} \label{sec:introduction}

Large language models (LLMs) are increasingly being adapted for resource-constrained edge devices \citep{qu2025mobile}, where they can enable adaptive networking, sensing, and cross-layer optimization \citep{shao2024wirelessllm} without relying on the cloud. 
Our previous work demonstrated that on-device LLMs can improve wireless performance by tuning cross-layer parameters in single-device settings, leveraging local context such as user location and time of day \citep{lee2025ondevicellm}. 
While promising, this prior line of research leaves open the question of how LLM-based optimization can extend beyond isolated devices to cooperative, device-to-device (D2D) scenarios.

In D2D communication, link performance depends not only on the local device state but also on peer conditions such as battery level. 
For example, when the publisher has ample energy but the subscriber is low on battery, an energy-preserving configuration may be preferable even at the cost of slightly higher latency. 
Designing a learning system that accounts for such asymmetric, two-sided context remains an open challenge. 
Moreover, efficient on-device deployment requires parameter-efficient fine-tuning (PEFT) strategies \citep{bucher2024fine, han2024peftsurvey}, but it is unclear how different PEFT modules compare in this setting, or how to ensure that training signals faithfully reflect heterogeneous application and device constraints.  

\begin{figure}[!h]
    \centering
    \includegraphics[width=0.6\linewidth]{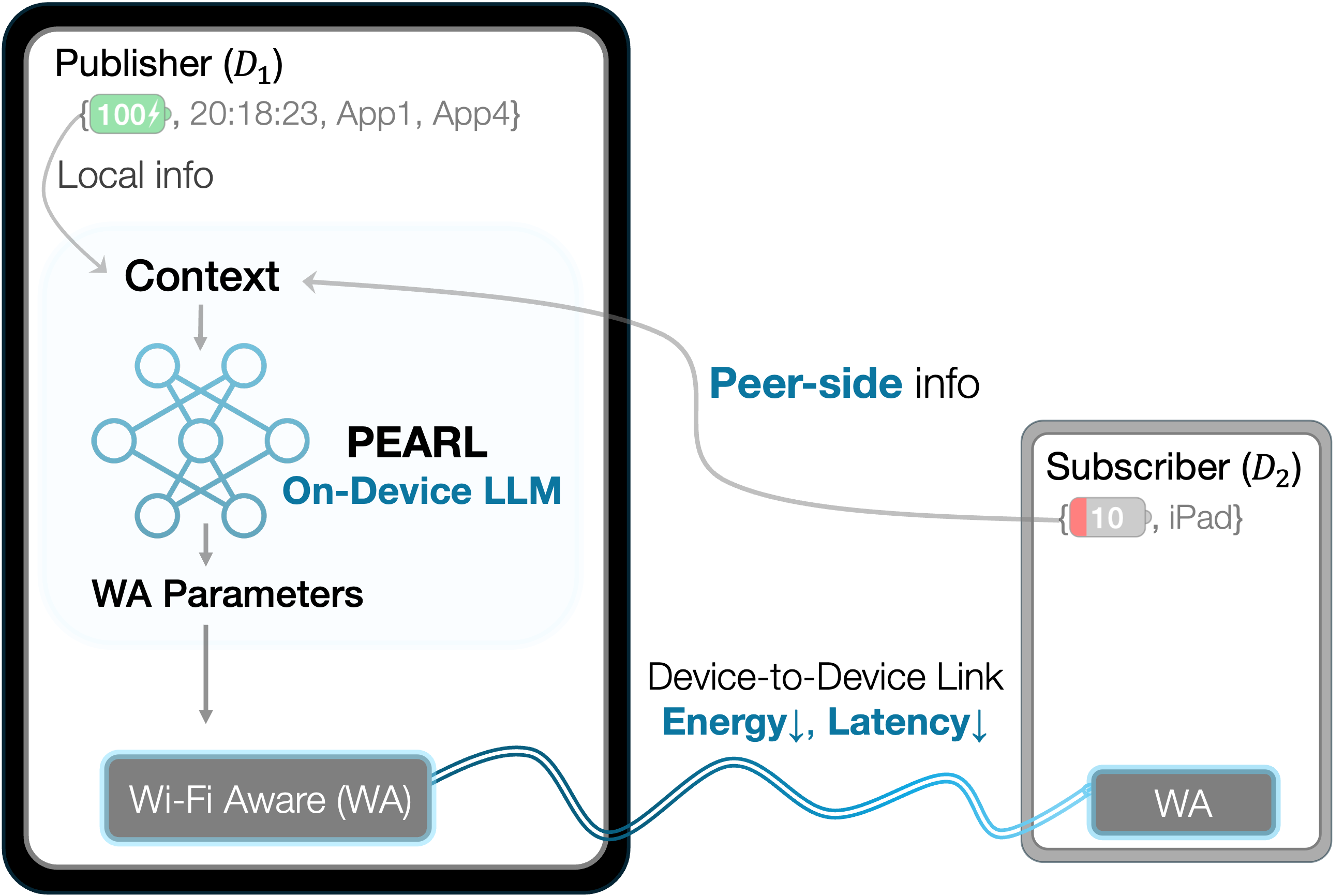}
    \caption{System overview of \textbf{PEARL}. The agent runs on the publisher ($D_1$), combining local context (\eg battery level, time, running applications) with peer-side context shared by the subscriber ($D_2$, \eg battery level, device type). Based on these inputs, PEARL selects WA parameters $(\texttt{PerformanceMode}, \texttt{AccessCategory})$, which configure the WA stack to minimize latency and energy on the D2D link.}
    \label{fig:system_overview}
\end{figure}



\BfPara{Contributions}  
We propose \textbf{PEARL} (\textit{Peer-Enhanced Adaptive Radio via On-Device LLM}), the first on-device LLM framework for cooperative cross-layer D2D optimization. 
Our key contributions are:  

(i) \textbf{Peer-context integration} (Sec.~\ref{sec:architecture}, \ref{sec:impact_peer_info}): subscriber-side information is incorporated into the model input, enabling two-sided adaptation beyond single-device optimization.  

(ii) \textbf{Context-aware reward} (Sec.~\ref{sec:reward}, \ref{sec:impact_reward_design}): a reward function that normalizes latency by application-specific tolerances and modulates energy by device battery state, providing richer supervision for KL-based fine-tuning.  

(iii) \textbf{Head-based PEFT} (Sec.~\ref{sec:post-training}, \ref{sec:impact_post_train_module}): lightweight classification heads (PEARL and PEARL-Lite) that achieve strong objective scores with sub-20\,ms inference, narrowing the gap to the oracle while remaining deployable on mobile hardware.

Extensive evaluation on \texttt{Wi-Fi Aware} (Sec.~\ref{sec:experiments} and Table~\ref{tab:non-llm}) shows that PEARL consistently outperforms heuristic baselines and compact non-LLM models, demonstrating that peer-aware, reward-aligned training substantially improves both efficiency and robustness in D2D optimization.

\section{Problem Formulation} \label{sec:problem}

\BfPara{Scenario}  
We consider a D2D link established via \texttt{Wi-Fi Aware} (WA) between a \emph{publisher} $D_1$ and a \emph{subscriber} $D_2$ \citep{campsmur2015wifi}. 
The publisher transmits application data while selecting cross-layer WA parameters that directly affect link performance. 
The decision model leverages both local and peer-side context, including device battery levels, device type (\eg iPhone, iPad), time of day, and foreground application type sampled from a synthetic, time-varying distribution.  

\BfPara{Action Space}  
At each decision step, the model chooses one of 8 discrete parameter tuples  
$(\texttt{PerformanceMode}, \texttt{AccessCategory})$, where  
$\texttt{PerformanceMode} \in \{\texttt{realtime}, \texttt{bulk}\}$ and  
$\texttt{AccessCategory} \in \{\texttt{bestEffort}, \texttt{background}, \texttt{interactiveVideo}, \texttt{interactiveVoice}\}$.  

\BfPara{Objective}  
The objective is to balance link latency and device energy consumption while satisfying application-specific requirements.
Both local (publisher) and peer (subscriber) context guide parameter selection. 
For example, when the subscriber has low battery, the model should prefer energy-preserving modes even at the cost of slightly higher latency.  
\section{Method} \label{sec:method}

\subsection{System Architecture} \label{sec:architecture}
PEARL builds on a pre-trained large language model (LLM) as a general-purpose feature encoder, and adapts it into a lightweight classifier for cross-layer parameter selection \citep{wu2024netllm}. 
Context features from both publisher and subscriber devices (\eg battery levels, time, application) are embedded into a structured prompt and passed to the frozen LLM backbone. 
A task-specific classification head is attached on top, predicting one of the discrete parameter tuples $(\texttt{PerformanceMode}, \texttt{AccessCategory})$.  

We consider two variants: PEARL (Head + Low-Rank Adaptation (LoRA) \citep{hu2022lora}), where the classification head is trained jointly with lightweight LoRA adapters inserted into the LLM, and PEARL-Lite (Head-only), where only the head is trained while the backbone remains frozen.
For comparison, we also evaluate \emph{LoRA-only}, where decisions are generated through the language modeling head without a classification head. 

\begin{figure}[!h]
    \centering
    \includegraphics[width=.99\columnwidth]{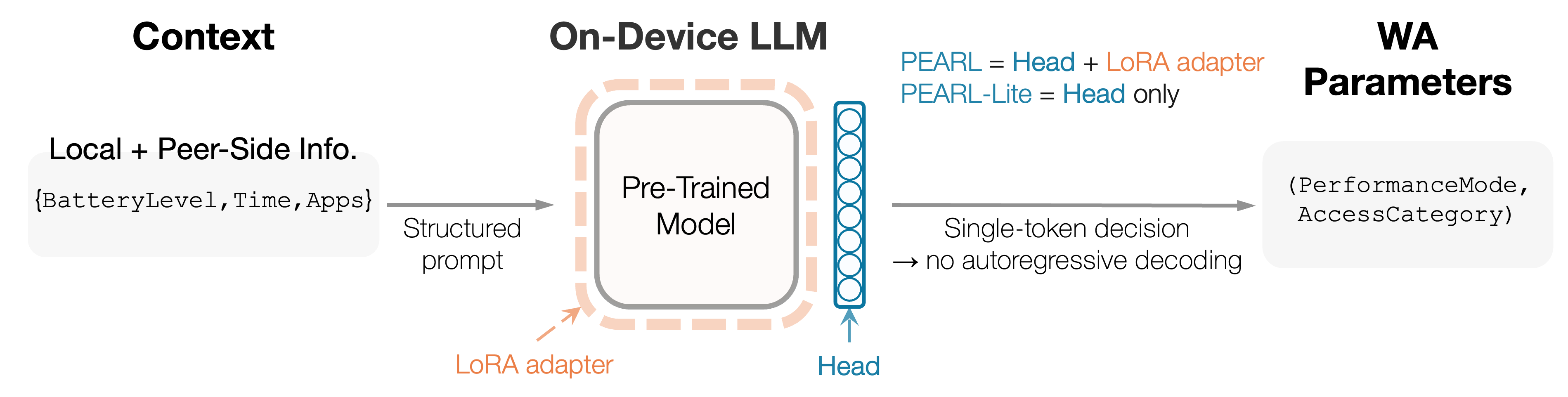}
    \caption{Architecture of PEARL. Context features are encoded into a structured prompt and passed through a frozen LLM encoder. In \textbf{PEARL-Lite}, a classification head directly predicts one of 8 WA parameter tuples. In \textbf{PEARL}, LoRA adapters are added to the encoder and trained jointly with the head. In both cases, the head produces a single-token decision, avoiding autoregressive decoding.}
    \label{fig:architecture}
\end{figure}

At inference, context features are encoded into a structured prompt, passed through the frozen LLM encoder, and mapped by the classification head to one of the 8 WA parameter tuples (Fig.~\ref{fig:architecture}).

\subsection{Context-Aware Reward Design} \label{sec:reward}

We design a reward function that captures both application-specific latency requirements and device battery states. 
Each is normalized into a \emph{score}, where higher values indicate better performance.  

For an application $a$ with latency tolerance $L_a$ and parameters $p$ yielding average latency $\ell(p)$, the latency score is
\[
R_{\text{lat}}(a,p) = \max\!\left(100 - 100 \cdot \tfrac{\ell(p)}{L_a},\, 0\right).
\]

For a device $d$ with battery level $b_d$ and energy usage $E(p)$ under parameters $p$, the energy score is
\[
R_{\text{eng}}(d,p) = \frac{b_d}{E(p)} .
\]

The overall reward combines these terms across active applications $\mathcal{A}$ and devices $\mathcal{D}$:
\[
R(p) = w_L \cdot \tfrac{1}{|\mathcal{A}|} \sum_{a \in \mathcal{A}} R_{\text{lat}}(a,p)
     - w_P \cdot \tfrac{1}{|\mathcal{D}|} \sum_{d \in \mathcal{D}} \tfrac{1}{R_{\text{eng}}(d,p)} .
\]

Here, $\ell(p)$ and $E(p)$ are the measured latency and energy under parameters $p$, 
$\mathcal{A}$ is the set of active applications, and $\mathcal{D}=\{\text{publisher}, \text{subscriber}\}$ the two devices considered. 
Weights $w_L$ and $w_P$ control the trade-off between latency and energy.  

This formulation produces a reward distribution that yields \emph{soft labels} for KL-based fine-tuning, preserving relative action preferences and improving generalization (Appendix~\ref{sec:metric}, \ref{sec:snapshot}).

\subsection{Post-Training Strategies} \label{sec:post-training}
We evaluate several post-training strategies for adapting the LLM encoder and prediction head: 

\begin{itemize}
    \item \textbf{Supervised Fine-Tuning (SFT):} We compare cross-entropy (CE) loss with Kullback–Leibler (KL) divergence loss. 
    CE uses hard argmax labels, while KL leverages the full reward distribution as soft labels, preserving relative preferences across actions \citep{hinton2015distilling}. 
    \item \textbf{Preference Optimization (DPO):} Optionally, we apply Direct Preference Optimization \citep{rafailov2023dpo} on paired contexts, where the higher-reward action is labeled as \emph{preferred} and the lowest-reward action as \emph{non-preferred}.
    \item \textbf{Head-based efficiency:} In PEARL and PEARL-Lite, the classification head consumes the last hidden state and outputs logits over the 8 actions, producing a one-token decision. 
    This avoids autoregressive decoding and yields sub-20 ms inference on edge hardware.
\end{itemize}

These strategies allow us to study the trade-offs between reward alignment, training cost, and inference efficiency (see Section~\ref{sec:experiments}).
\section{Experiments} \label{sec:experiments}

\subsection{Setup}
\BfPara{Dataset}  
We construct a simulation dataset from WA measurements combined with synthetic application usage distributions. 
For each parameter tuple $(\texttt{PerformanceMode}, \texttt{AccessCategory})$, we measure average latency and power usage during controlled sessions. 
Each sample consists of $(\text{context}, \text{action}, \text{score})$, with scores computed as described in Section~\ref{sec:reward}. 
Further details and parameter settings are provided in Table~\ref{tab:setup} and Appendix~\ref{sec:dataset}.

\BfPara{Baselines}  
We evaluate our proposed framework for context-driven cross-layer optimization under the \textbf{PEARL} and \textbf{PEARL-Lite} variants introduced in Section~\ref{sec:architecture}. 
For comparison, we also include: (i) an \emph{oracle}, which selects the configuration that maximizes the context-aware reward derived from ground-truth latency and energy — effectively representing the optimal balance between latency and energy consumption under the defined application constraints; 
(ii) \emph{LoRA-only} (\textbf{LoRA}); (iii) a rule-based baseline (Rule), which selects WA parameters heuristically based only on application type (see Table~\ref{tab:preferred} in Appendix~\ref{sec:baseline}); and 
(iv) two fixed baselines, \emph{Fixed (RT/IV)} (\textbf{fix-RT/IV}, $(\texttt{realtime}, \texttt{interactiveVoice})$) and \emph{Fixed (Bulk/BG)} (\textbf{fix-Bulk/BG}, $(\texttt{bulk}, \texttt{background})$).

\BfPara{Metrics}  
We report three metrics aligned with the definitions in Section~\ref{sec:reward}:  
(i) the \textbf{Objective Score}, corresponding to the overall reward $R(p)$ that combines normalized latency and energy;  
(ii) \textbf{Latency}, the latency score $R_{\text{lat}}$ normalized by application-specific tolerance (higher is better); and  
(iii) \textbf{Energy}, the energy score $R_{\text{eng}}$, which reflects normalized energy consumption relative to device battery state (higher is better).  
Unless otherwise specified, results are averaged across scenarios covering different times of day and battery levels.

\subsection{Performance Comparison}
\begin{figure}[!h]
\centering
\begin{subfigure}[b]{0.32\textwidth}
    \includegraphics[width=\textwidth]{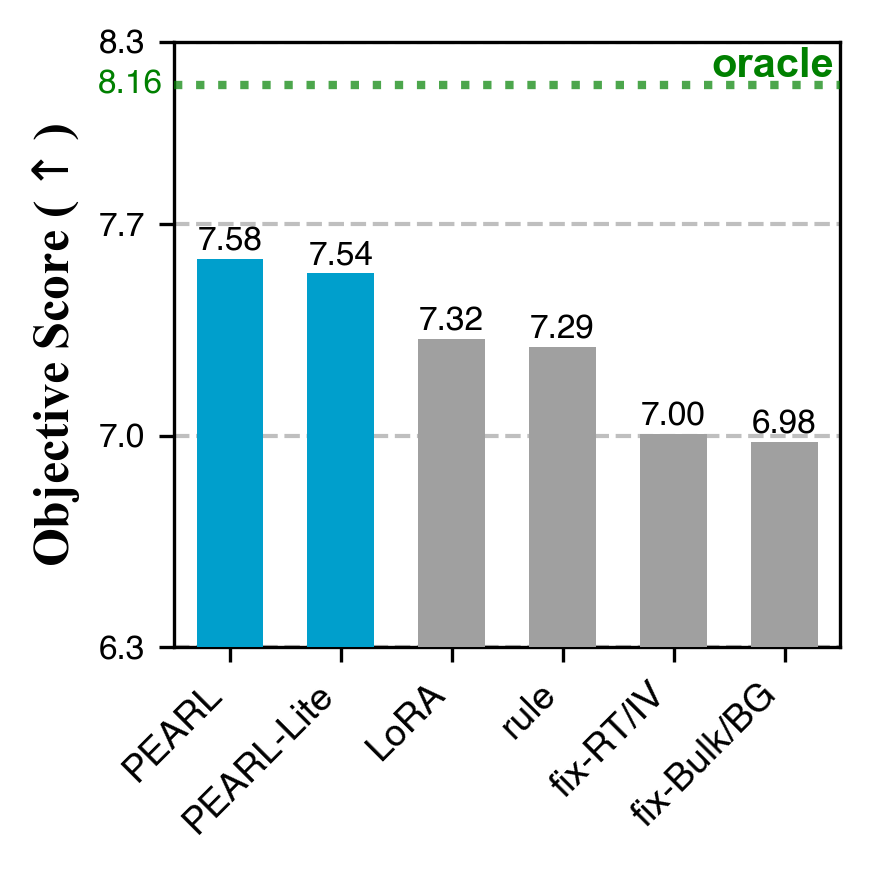}
\end{subfigure}
\hfill
\begin{subfigure}[b]{0.32\textwidth}
    \includegraphics[width=\textwidth]{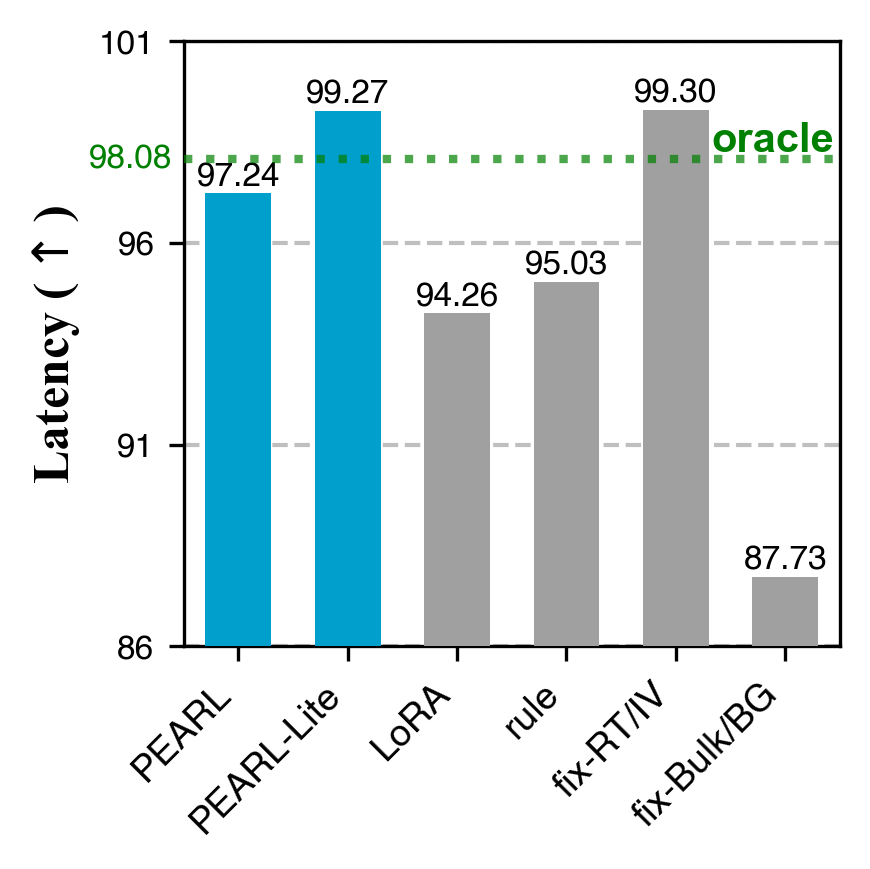}
\end{subfigure}
\hfill
\begin{subfigure}[b]{0.32\textwidth}
    \includegraphics[width=\textwidth]{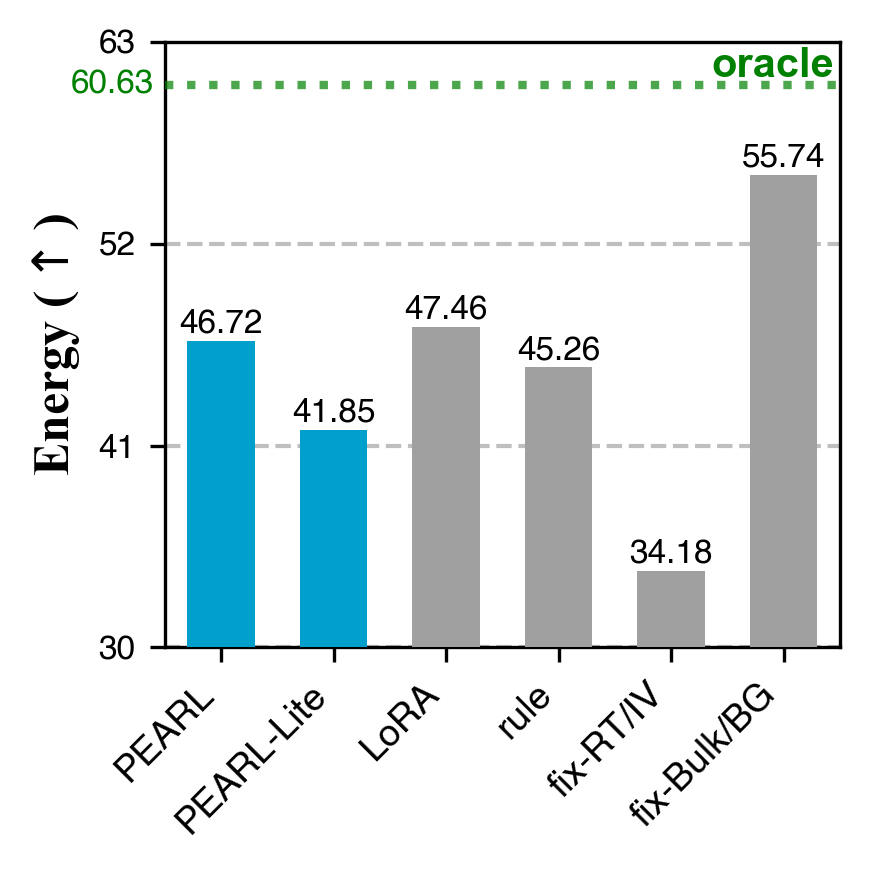}
\end{subfigure}
\caption{Objective score, latency, and energy comparison across PEARL variants and baselines.}
\label{fig:performance_comparison}
\end{figure}

Figure~\ref{fig:performance_comparison} shows that \textbf{PEARL} outperforms rule-based and fixed baselines on the joint objective score. 
The default \textbf{PEARL} variant achieves $7.58$, with \textbf{PEARL-Lite} close behind at $7.54$; both are about $7\%$ below the oracle ($8.16$) but remain well above \textbf{LoRA-only} ($7.32$) and \textbf{Rule} ($7.29$). 
While \textbf{PEARL-Lite} emphasizes lower latency at the cost of higher energy, \textbf{PEARL} strikes a more balanced trade-off, leading to the best overall performance. 
These results demonstrate that head-based designs enable effective context-driven adaptation and substantially narrow the gap to the oracle compared to heuristic strategies.

\subsection{Impact of Peer Context Information} \label{sec:impact_peer_info}
\begin{table}[!h]   
\centering
\resizebox{\columnwidth}{!}{
\centering
\begin{tabular}{l l|ccc|ccc}
\toprule
\multirow{2}{*}{\textbf{Variant}} & & \multicolumn{3}{c|}{\textbf{Aggregated Scenario}} & \multicolumn{3}{c}{\textbf{Cooperative Scenario (Low-Battery Subscriber)}} \\
\cmidrule(lr){3-6}\cmidrule(lr){6-8}
& & \textbf{Objective Score} $\uparrow$ & \textbf{Latency} $\uparrow$ & \textbf{Energy} $\uparrow$ & \textbf{Objective Score} $\uparrow$ & \textbf{Latency} $\uparrow$ & \textbf{Energy} $\uparrow$ \\
\midrule
w/ peer info & \multirow{2}{*}{(PEARL)} & 7.58 & 97.24 & 46.72 & 5.38 & 94.55 & 24.56 \\
w/o peer info & & 7.55 & 97.98 & 44.50 & 5.04 & 99.13 & 20.53 \\
\midrule[0.2pt]
w/ peer info & \multirow{2}{*}{(PEARL-Lite)}      & 7.54 & 99.27 & 41.85 & 5.22 & 99.14 & 21.29 \\
w/o peer info & & 7.49 & 99.27 & 40.99 & 5.03 & 99.13 & 20.47 \\
\bottomrule
\end{tabular}}
\vspace{0.5em}
\caption{Impact of including peer information in the context. In the aggregated scenario, 
models trained \emph{w/ peer info} achieve consistently higher objective scores than those \emph{w/o peer info}. 
In the cooperative scenario—where the subscriber is low on battery and the publisher adapts its configuration to assist—adding peer info reduces energy consumption by up to $\sim$16\% while maintaining reasonable latency, highlighting where peer awareness provides the greatest benefit.
}
\label{tab:peer_info}
\end{table} 

Table~\ref{tab:peer_info} shows that incorporating peer (subscriber) information consistently improves performance. 
In aggregated scenarios, PEARL and PEARL-Lite \emph{w/ peer info} achieve slightly higher objective scores than their \emph{w/o peer info} counterparts. 
In the cooperative case, \textbf{PEARL} \emph{w/ peer info} reduces energy consumption by $\sim$16\% while maintaining latency within $0.5\%$ of \emph{w/o peer info}, confirming that peer awareness provides tangible benefits in asymmetric conditions where coordination is critical. Additional qualitative snapshots illustrating this effect are provided in Appendix~\ref{sec:snapshot}.

\subsection{Impact of Context-Aware Reward Design} \label{sec:impact_reward_design}
\begin{minipage}{0.7\textwidth}
Figure~\ref{fig:reward_design} shows that context-aware reward improves performance for both 
PEARL and PEARL-Lite. By normalizing latency against application-specific 
tolerances and incorporating device battery states into the energy term, the design generates 
richer supervision signals that better reflect heterogeneous requirements. 
Although absolute gains over the naive formulation are modest, they consistently yield higher 
objective scores, indicating more accurate adaptation. 
This demonstrates that careful reward design is crucial for leveraging context effectively in cross-layer optimization and for ensuring robustness across diverse usage scenarios.
\end{minipage}
\hfill
\begin{minipage}{0.28\textwidth}
    \centering
    \captionsetup{font=footnotesize}
    \includegraphics[width=.9\linewidth]{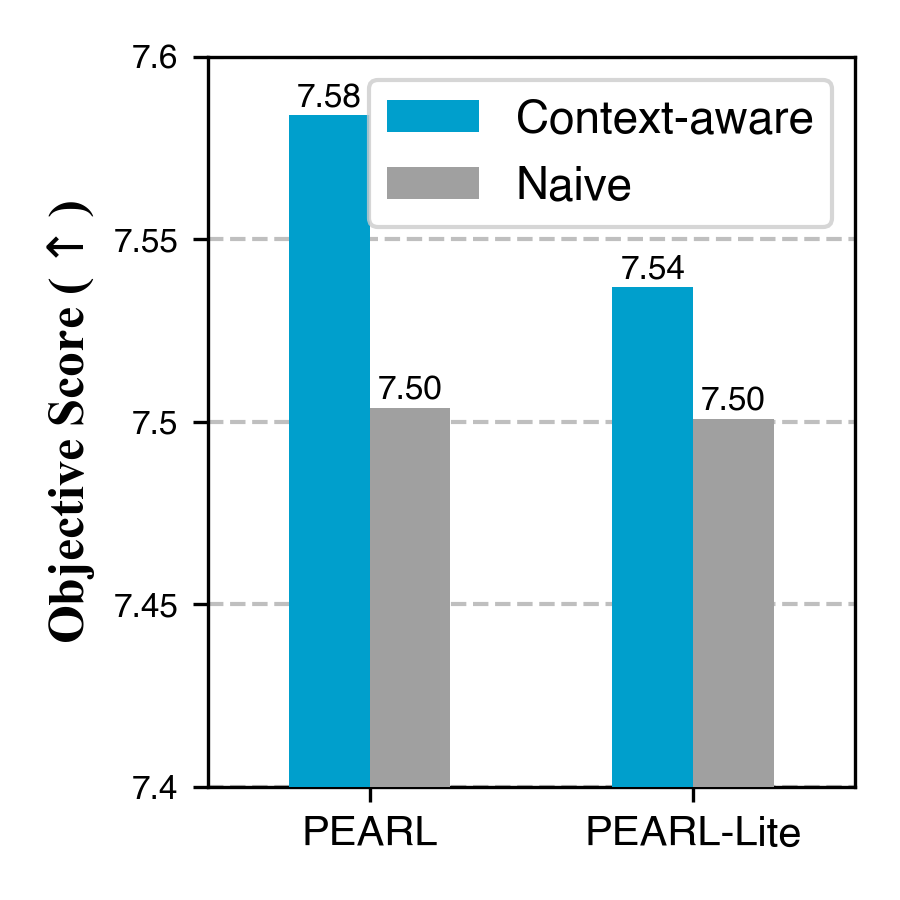}
    \vspace{-1.em}
    \captionof{figure}{Objective scores with context-aware vs.\ naive reward.}
    \label{fig:reward_design}
\end{minipage}

\subsection{Effect of Post-Training Strategy}
\begin{figure}[!h]
\centering
\begin{subfigure}[b]{0.45\textwidth}
    \includegraphics[width=\textwidth]{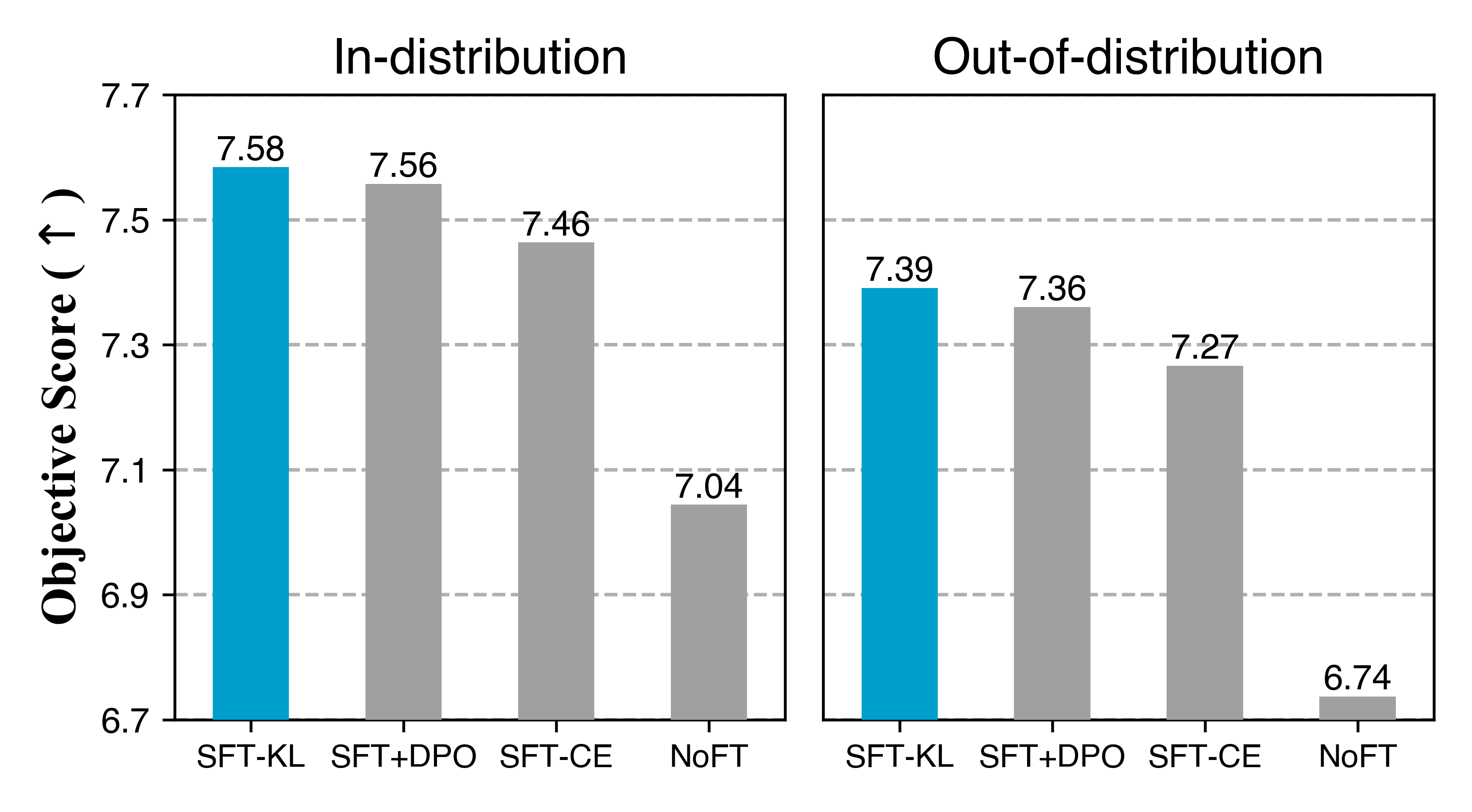}
    \caption{\textbf{PEARL}}
\end{subfigure}
\hfill
\begin{subfigure}[b]{0.45\textwidth}
    \includegraphics[width=\textwidth]{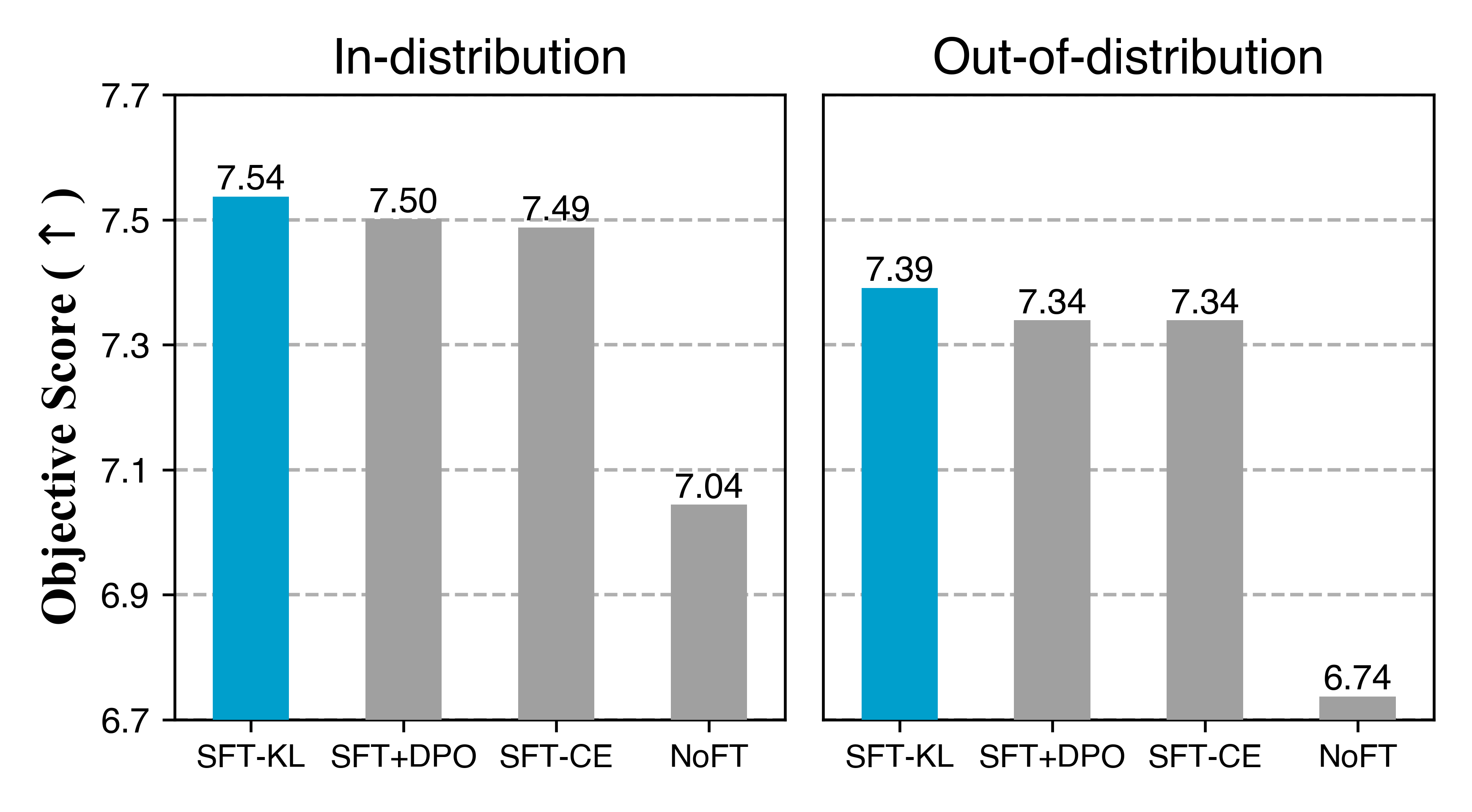}
    \caption{\textbf{PEARL-Lite}}
\end{subfigure}
\caption{Objective scores for different training strategies. KL-based fine-tuning consistently outperforms CE and DPO, both in-distribution and out-of-distribution (OOD).}
\label{fig:training_strategy}
\end{figure}

Figure~\ref{fig:training_strategy} compares post-training strategies for \textbf{PEARL} and \textbf{PEARL-Lite}. 
Without fine-tuning (NoFT), objective scores are substantially lower in both in-distribution and OOD cases. 
Supervised fine-tuning with cross-entropy (SFT-CE) improves scores, while adding preference optimization (SFT+DPO) yields only marginal gains despite higher cost. 
In contrast, KL-based fine-tuning (SFT-KL) consistently achieves the best scores across both architectures. 
Its advantage stems from using soft labels derived from reward distributions, which preserve relative action preferences instead of collapsing them into hard labels. 
This provides richer supervision signals, enabling more nuanced trade-offs and better generalization across diverse scenarios.

\subsection{Efficiency of Post-Training Modules} \label{sec:impact_post_train_module}
\begin{table}[!h]   
\centering
\resizebox{.95\columnwidth}{!}{
\centering
\begin{tabular}{l | c | r c c | c c}
\toprule
\multirow{2}{*}{\textbf{Baseline}} & 
\multirow{2}{*}{\textbf{Objective Score} $\uparrow$} & 
\multicolumn{3}{c|}{\textbf{Inference} $\downarrow$} & 
\multicolumn{2}{c}{\textbf{Training} $\downarrow$} \\
\cmidrule(lr){3-5} \cmidrule(lr){6-7}
 &  & \textbf{Time (ms)} & \textbf{Tokens} & \textbf{RAM (GB)} & \textbf{RAM (GB)} & \textbf{Time (min/epoch)} \\
\midrule
Base          & 7.04 & 760.7 (×1.0)   & 48.9 & 6.58 & —     & —  \\
LoRA-only     & 7.32 & 243.1 (×3.1)   & 7.62 & 9.39 & 29.17 & 30 \\
\textbf{PEARL-Lite}    & 7.54 & \textbf{14.5 (×52)} & 1.00 & 6.80 & 11.03 & 12 \\
\textbf{PEARL}         & \textbf{7.58} & 40.3 (×18.9) & 1.00 & 9.44 & 46.83 & 38 \\
\bottomrule
\end{tabular}

}
\vspace{0.2em}
\caption{Performance and efficiency of post-training modules. Parentheses denote speedup vs.\ Base (NoFT) . 
“Tokens” = average generated tokens per decision; head-based models always produce 1.00 (classification head). 
\textbf{PEARL-Lite} is most efficient, while \textbf{PEARL} yields the highest objective score.}
\label{tab:efficiency}
\end{table} 

Table~\ref{tab:efficiency} highlights the trade-off between performance and efficiency across different
post-training modules. \textbf{PEARL} achieves the highest objective score ($7.58$) but incurs moderate
training cost, while \textbf{PEARL-Lite} attains nearly comparable performance ($7.54$) with substantially
lower inference latency ($14.5$ ms, $52\times$ faster than the Base) and reduced training memory. 
In contrast, \textbf{LoRA-only} yields lower objective scores ($7.32$) and slower inference, underscoring 
that head-based designs provide the most favorable balance for on-device deployment. 
This efficiency comes from the classification head: it consumes the last token’s hidden state to produce logits 
over the 8-action space (a single token), whereas LoRA-only relies on text generation, producing $\sim$8 tokens on average. 
\section{Discussion}\label{sec:discussion}
\BfPara{Q1 — Interpretation: \textit{How do peer context and a context-aware reward improve PEARL’s on-device D2D decisions?}}  
Incorporating subscriber-side context reduces partial observability and enables coordinated adaptation across devices. 
This leads to consistently higher objective scores, and in cooperative cases (\eg low-battery subscriber) reduces energy by $\sim$16\% with negligible latency change ($\leq$0.5\%) (Table~\ref{tab:peer_info}), showing that peer awareness makes PEARL more responsive to asymmetric conditions.  
The context-aware reward amplifies this effect by aligning training with heterogeneous constraints: latency is normalized by application-specific tolerances, and energy is scaled according to device battery states (Sec.~\ref{sec:reward}, Fig.~\ref{fig:reward_design}, Appendix~\ref{sec:metric}).
Together, these choices guide PEARL toward energy-efficient yet latency-safe actions, improving robustness in asymmetric scenarios.

\BfPara{Q2 — Practical Implications: \textit{How does PEARL’s head-based design enable efficient deployment on real edge devices?}}  
PEARL’s head-based design performs inference by emitting a \emph{single} classification token from the last hidden state, avoiding autoregressive decoding and its latency/memory overhead. 
This allows \textbf{PEARL-Lite} to run in under 20\,ms ($\sim$52$\times$ faster than the base) with low memory usage, while full \textbf{PEARL} achieves the highest objective score with roughly 3× inference latency and 3–4× training memory relative to PEARL-Lite — a trade-off that yields measurable gains in accuracy and robustness.
By contrast, LoRA-only is slower and less effective due to multi-token generation. 
In deployment, \textbf{PEARL-Lite} is well-suited when latency or GPU memory (VRAM) budgets are tight, \textbf{PEARL} is preferable when slight overhead is acceptable for maximum accuracy, and merging adapters into the base can approximate PEARL’s performance at near PEARL-Lite cost (Table~\ref{tab:efficiency}). 
Overall, these properties give PEARL the latency–energy–memory profile required for always-on, on-device cross-layer control without server support.

\BfPara{Q3 — Alternatives: \textit{Why not replace PEARL with a compact feed-forward network (FFN/MLP), a reinforcement-learning (RL) baseline, or a lightweight Transformer encoder?}} 
While small feed-forward networks or RL baselines can be efficient for narrow tasks, they struggle with mixed-format context (numeric + categorical) and typically require additional feature engineering. 
Lightweight Transformer encoders such as \texttt{MiniLM} (22.7M parameters) \citep{wang2020minilm} can process heterogeneous inputs more flexibly and achieve very low inference latency ($\sim$5–6\,ms). 
However, as Table~\ref{tab:non-llm} shows, they yield lower objective and OOD scores than PEARL. 
By contrast, PEARL and PEARL-Lite leverage pretrained language modeling to naturally parse diverse context, providing stronger generalization while still running in the sub-20\,ms range. 
Thus, although compact models may be attractive under extreme latency budgets, PEARL offers the best balance of accuracy, robustness, and deployability for always-on D2D optimization.

\begin{table}[!h]   
\centering
\resizebox{.8\columnwidth}{!}{
\centering
\begin{tabular}{lccc}
\toprule
\textbf{Baseline} & \textbf{Objective Score} $\uparrow$ & \textbf{OOD Score} $\uparrow$ & \textbf{Inference Time (ms)} $\downarrow$ \\
\midrule
\textbf{PEARL}         & \textbf{7.58} & \textbf{7.39} & 40.3 \\
\textbf{PEARL-Lite}    & 7.54 & \textbf{7.39} & 14.5 \\
\midrule
MiniLM (Head+LoRA)     & 7.51 & 7.34 & 6.0 \\
MiniLM (Head)          & 7.49 & 7.34 & \textbf{5.1} \\
\bottomrule
\end{tabular}

}
\vspace{0.2em}
\caption{Comparison of PEARL with a lightweight Transformer encoder (\texttt{all-MiniLM-L6-v2}, 22.7M).}
\label{tab:non-llm}
\end{table}

\BfPara{Q4 — Limitations \& Future Directions}  
We provide a proof-of-concept demo, with code, dataset, and a full demonstration video available at \url{https://github.com/abman23/pearl} to ensure reproducibility. 
Nevertheless, rigorous validation on hardware testbeds with real-time counters (\eg per-packet latency and energy) is required for robust evaluation.
Our study is currently limited to a narrow task space (2 $\texttt{PerformanceMode}$ options $\times$ 4 $\texttt{AccessCategory}$ options) and a single protocol (WA).
Extending PEARL to richer parameter spaces, additional protocols (\eg 5G), and heterogeneous devices remains an open challenge.
Moreover, while head-based inference provides fast and accurate decisions, scaling to multiple tasks with separate heads may become burdensome.
Future work should explore multi-task optimization strategies that preserve efficiency while maintaining accuracy.

\newpage
\bibliographystyle{plainnat}
\bibliography{refs}

\newpage
\appendix
\section{Real-World Demonstration}\label{sec:demo}

To validate the feasibility of \textbf{PEARL} beyond simulation, we implemented a real-world prototype as an \texttt{iPadOS} app. 
The app establishes D2D communication between two devices using Apple’s WA framework and integrates Apple’s on-device Foundation Models (AFM) to run the LLM agent directly on the publisher \citep{apple2024foundation}. 
The agent monitors local and peer-side context and dynamically updates WA parameters $(\texttt{PerformanceMode}, \texttt{AccessCategory})$ to reconfigure the ongoing connection. 
We deploy the app on two iPad Pro devices (Apple M4 chip, 8~GB RAM, \texttt{iPadOS}~26.0). 
A full demonstration video is available at \url{https://github.com/abman23/pearl}.

\begin{figure}[!h]
\centering
\includegraphics[width=.65\textwidth]{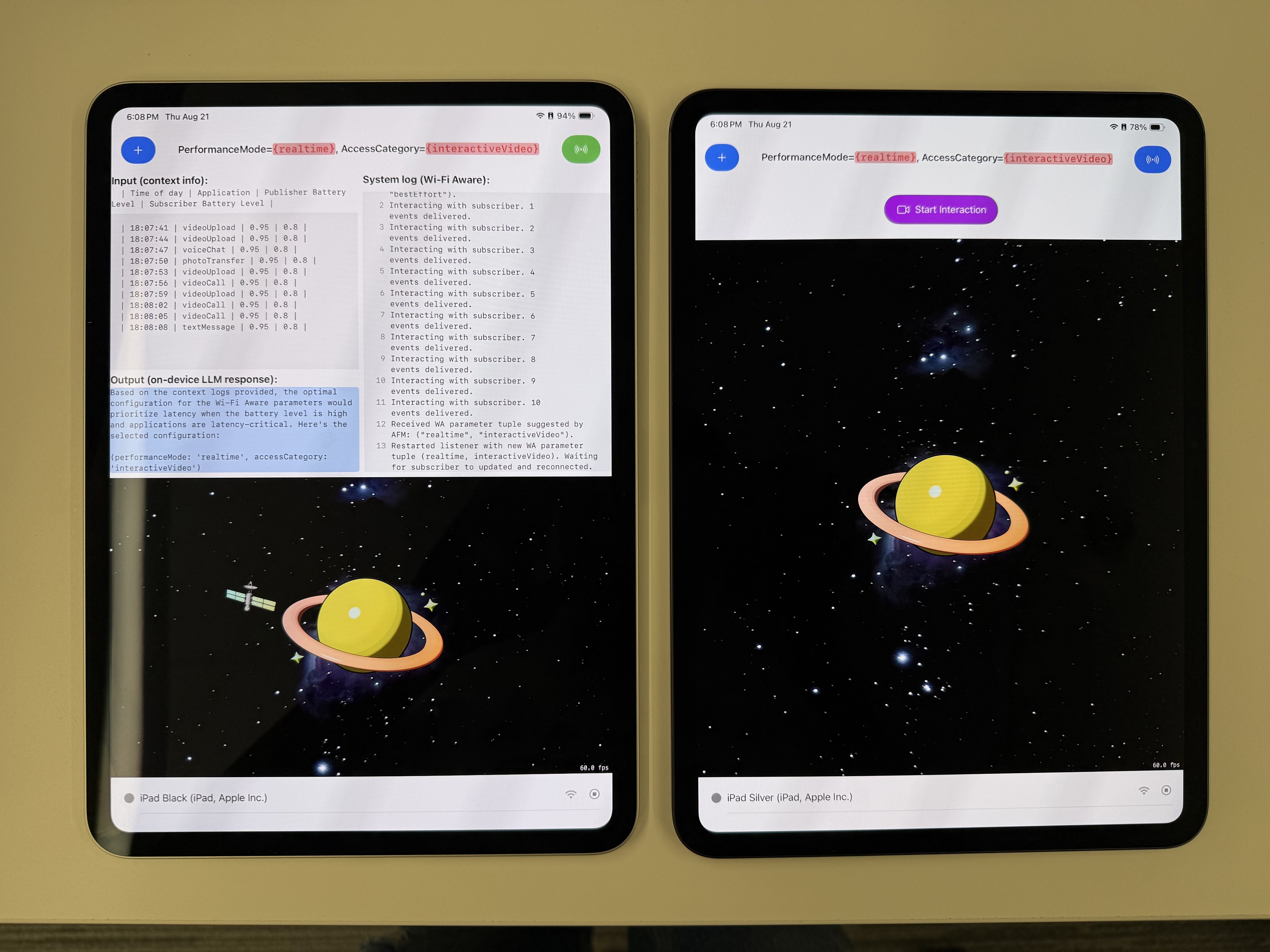}
\caption{Prototype demonstration of PEARL on two iPad Pro devices. The publisher (left) hosts the on-device LLM agent, while the subscriber (right) receives updated WA parameters.}
\label{fig:demo-realworld}
\end{figure}

\begin{figure}[!h]
\centering
\includegraphics[width=.85\textwidth]{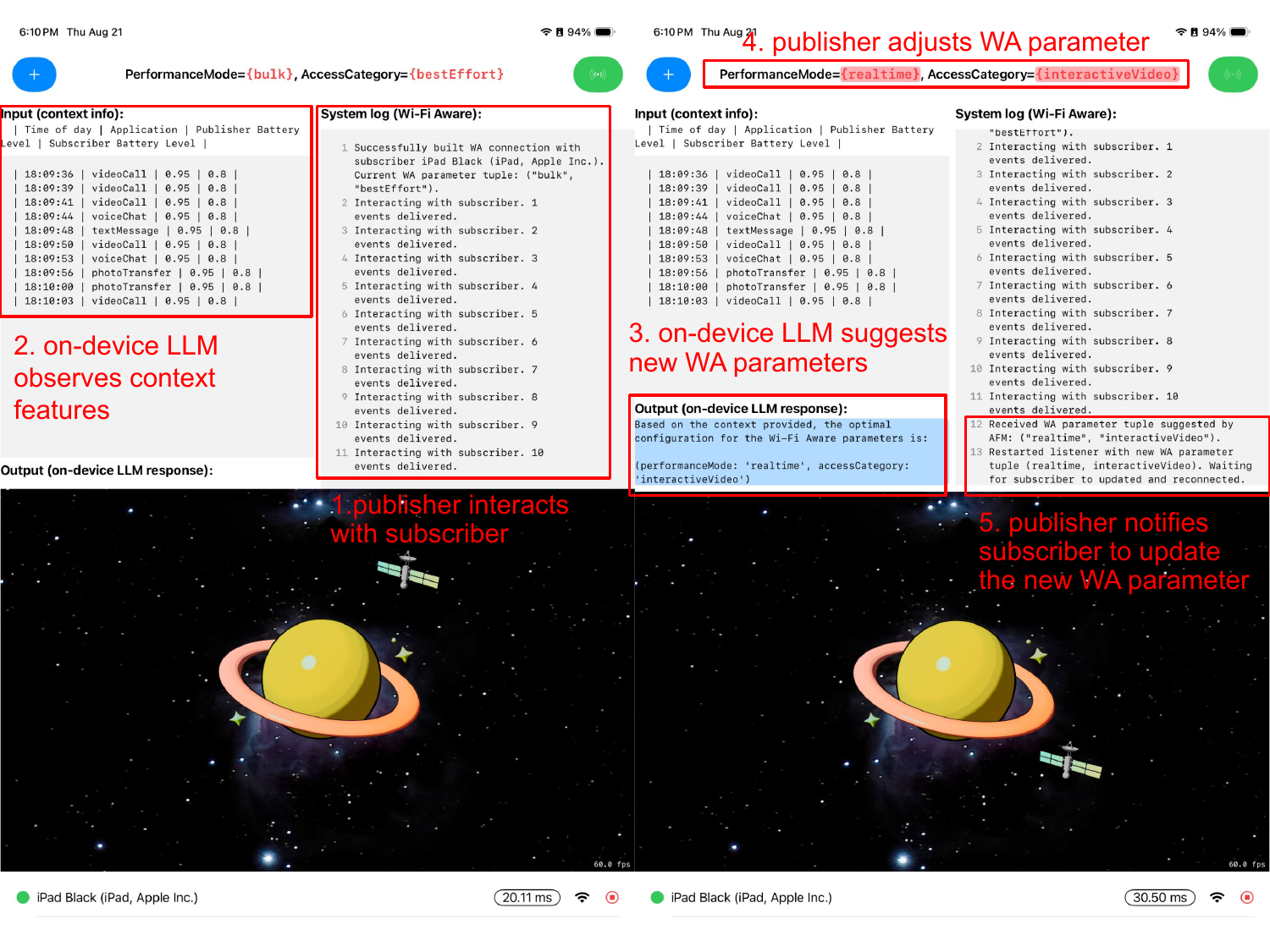}
\caption{Step-by-step illustration of WA parameter tuning. The publisher observes context features and proposes a new configuration $(\texttt{realtime}, \texttt{interactiveVideo})$, then reconfigures the WA link and notifies the subscriber to update.}
\label{fig:demo-steps}
\end{figure}

\newpage
\section{Setup}\label{sec:setup}

Table~\ref{tab:setup} summarizes the hardware, software, dataset, and training configurations. 
For task design, we synthesize eight application types with time-of-day usage distributions that mimic typical mobile usage patterns (Fig.~\ref{fig:in-dsitribution}). 
These distributions are used to sample application types as part of the context features in training and in-distribution evaluation. 
To test generalization, we construct an alternative distribution (Fig.~\ref{fig:ood}) to generate out-of-distribution (OOD) evaluation sets. 

\begin{table}[!h]   
\centering
\resizebox{.9\columnwidth}{!}{
\centering
\begin{tabular}{l p{11cm}}
\toprule[1pt]
\textbf{HW / SW} & \\
\midrule
Device (Publisher / Subscriber) & iPad Pro (M4, 8\,GB RAM) \\
OS & \texttt{iPadOS 26.0} \\
Frameworks & \texttt{Wi-Fi Aware}, \texttt{Foundation Models} \\
LLM Backbone & \texttt{Llama-3.2-3B}, \texttt{AFM‑on‑device} \citep{apple2024foundation} \\
\midrule
\textbf{Dataset (WA Log)} & \\
\midrule
Total Samples & 32,000 \\
Train / Test Split & 80\% / 20\% \\
Sampling Interval & 5\,s \\
\midrule
\textbf{Task Configuration} & \\
\midrule
Action Space & 2 modes × 4 categories = 8 actions \\
Scenarios & 16 combinations of time-of-day (morning, afternoon, evening, night) 
and device battery levels (both high, medium, low, publisher high / subscriber low) \\
Application Types & \texttt{textMessage}, \texttt{voiceChat}, \texttt{videoCall}, 
\texttt{sensorSync}, \texttt{photoTransfer}, \texttt{videoUpload}, 
\texttt{firmwareUpdate}, \texttt{mapSync} \\
Contextual Inputs & Time, App Types, Publisher Battery Level, Subscriber Battery Level \\
Reward Weights & $w_L = 0.1$, $w_P = 1.0$ \\
History Window & 10 past time steps \\
\midrule
\textbf{Training Configuration} & \\
\midrule
Learning Rate & $1\times 10^{-5}$ \\
Batch Size / Grad. Accum. & 4 / 16 (effective batch = 64) \\
Epochs & 5 \\
Weight Decay & 0.01 \\
Loss Functions & CE, KL, DPO \\
LoRA Rank / Alpha & 128 / 128 \\
DPO Beta & 0.1 \\
Optimizer & \texttt{AdamW} \\
GPU & 1 × NVIDIA L40S \\
\bottomrule[1pt]
\end{tabular}

}
\vspace{0.3em}
\caption{Experiment setup including hardware, dataset, task configuration, and training details.}
\label{tab:setup}
\end{table} 

\begin{figure}[!h]
\centering
\begin{subfigure}[b]{0.3\textwidth}
\begin{lstlisting}[basicstyle=\ttfamily\scriptsize]
morning:
    textMessage: 0.25,
    voiceChat: 0.20,
    videoCall: 0.25,
    mapSync: 0.20,
    photoTransfer: 0.10
afternoon:
    textMessage: 0.20,
    voiceChat: 0.20,
    videoCall: 0.25,
    sensorSync: 0.20,
    photoTransfer: 0.15
evening:
    textMessage: 0.10,
    voiceChat: 0.20,
    videoUpload: 0.30,
    photoTransfer: 0.20,
    videoCall: 0.20
night:
    firmwareUpdate: 0.40,
    sensorSync: 0.30,
    textMessage: 0.10,
    photoTransfer: 0.10,
    mapSync: 0.10
\end{lstlisting}
\caption{Training / in-distribution}
\label{fig:in-dsitribution}
\end{subfigure}
\hspace{2em}
\begin{subfigure}[b]{0.3\textwidth}
\begin{lstlisting}[basicstyle=\ttfamily\scriptsize]
morning:
    videoCall: 0.30,
    textMessage: 0.25,
    voiceChat: 0.20,
    photoTransfer: 0.15,
    videoUpload: 0.10
afternoon:
    videoCall: 0.25,
    textMessage: 0.25,
    voiceChat: 0.20,
    photoTransfer: 0.20,
    videoUpload: 0.10
evening: 
    videoUpload: 0.35,
    videoCall: 0.25,
    photoTransfer: 0.20,
    voiceChat: 0.15,
    textMessage: 0.05
night: 
    videoUpload: 0.30,
    videoCall: 0.25,
    textMessage: 0.20,
    voiceChat: 0.15,
    photoTransfer: 0.10
\end{lstlisting}
\caption{Out-of-distribution evaluation}
\label{fig:ood}
\end{subfigure}
\vspace{0.2em}
\caption{Synthetic application usage distributions used to generate context logs.}
\label{fig:app-distribution}
\end{figure}

\newpage
\section{Context-Aware Evaluation Metrics}\label{sec:metric}

We define two context-dependent metrics, \textbf{Latency} and \textbf{Energy}, to evaluate link quality under D2D dynamics. 
Unlike raw values, these metrics incorporate application- and device-specific constraints, providing not only quantitative measurement but also qualitative insight into how well system behavior aligns with user needs.  

\textbf{Latency} is computed as a function of WA parameters and application type, expressed as a percentage score indicating how well the observed link latency satisfies the \emph{latency tolerance} of the active application (Table~\ref{tab:latency-tolerance}).  
\textbf{Energy} is computed as a function of WA parameters and device battery level, reflecting how well the energy consumption of a configuration aligns with the current battery status.  

\begin{table}[!h]   
\centering
\resizebox{.42\columnwidth}{!}{
\centering
\begin{tabular}{lc}
\toprule
\textbf{Application Type} & \textbf{Latency Tolerance (ms)} \\
\midrule
\texttt{textMessage}      & 200 \\
\texttt{voiceChat}        & 50 \\
\texttt{videoCall}        & 100 \\
\texttt{sensorSync}       & 1,000 \\
\texttt{photoTransfer}    & 2,000 \\
\texttt{videoUpload}      & 5,000 \\
\texttt{firmwareUpdate}   & 10,000 \\
\texttt{mapSync}          & 500 \\
\bottomrule
\end{tabular}}
\vspace{0.3em}
\caption{Latency tolerance thresholds for each application type.}
\label{tab:latency-tolerance}
\end{table} 

To highlight the benefit of these metrics, we contrast them with raw measurements. 
Table~\ref{tab:latency-scores} shows that while raw latency (ms) varies significantly across time periods, the \textbf{Latency} score remains consistent with application tolerance. 
For example, although average link latency at night (15.61\,ms) is much higher than in the morning (4.97\,ms), most nighttime applications are delay-tolerant, producing similarly high scores.  
Likewise, Table~\ref{tab:energy-scores} compares \textbf{Energy} with raw energy consumption across high, medium, and low battery states. 
Here, the metric magnifies differences under low-battery conditions, assigning greater penalty and dynamically shifting the optimization objective toward energy efficiency.

\begin{table}[!h]   
\centering
\resizebox{.65\columnwidth}{!}{
\centering
\begin{tabular}{lcc}
\toprule
\textbf{Scenario (time, battery)} & \textbf{Latency Score $\uparrow$} & \textbf{Link Latency (ms) $\downarrow$} \\
\midrule
(morning, both medium)   & 98.41 & 4.97 \\
(afternoon, both medium) & 98.43 & 6.60 \\
(evening, both medium)   & 98.94 & 9.56 \\
(night, both medium)     & 98.95 & 15.61 \\
\bottomrule
\end{tabular}
}
\vspace{0.3em}
\caption{Comparison of context-aware \textbf{Latency} score and raw link latency across scenarios.}
\label{tab:latency-scores}
\end{table} 

\begin{table}[!h]   
\centering
\resizebox{.75\columnwidth}{!}{
\centering
\begin{tabular}{lcc}
\toprule
\textbf{Scenario (time, battery)} & \textbf{Energy Score $\uparrow$} & \textbf{Energy Consumption (\%/h) $\downarrow$} \\
\midrule
(afternoon, both high)   & 102.99 & 3.15 \\
(afternoon, both medium) & 59.46  & 3.24 \\
(afternoon, both low)    & 24.26  & 2.90 \\
\bottomrule
\end{tabular}
}
\vspace{0.3em}
\caption{Comparison of context-aware \textbf{Energy} score and raw energy consumption across scenarios.}
\label{tab:energy-scores}
\end{table}

\newpage
\section{Data Collection}\label{sec:dataset}


The simulation dataset is constructed from real WA logs, combined with synthetic application usages defined  in Fig.~\ref{fig:app-distribution}. To collect logs, we developed an \texttt{iPadOS} app that establishes a WA connection between two devices, simulates their interactions, and records context information. The app was deployed on two iPad Pro devices (Table~\ref{tab:setup}), where we executed 16 sessions of data collection, each consisting of 2,000 consecutive logs. Each session corresponds to one specific scenario defined by time of day and device battery levels. Battery levels were sampled uniformly across high, medium, and low states to ensure balanced representation of energy conditions.


During each session, we iterated over all WA parameter tuples in the action space so that every configuration $(\texttt{PerformanceMode}, \texttt{AccessCategory})$ was exercised. 
For each tuple, the app recorded both link latency values and battery usage, which are later used to compute the objective score as described in Section~\ref{sec:reward} and Section~\ref{sec:metric}.  

\begin{figure}[!h]
\centering
\includegraphics[width=.55\textwidth]{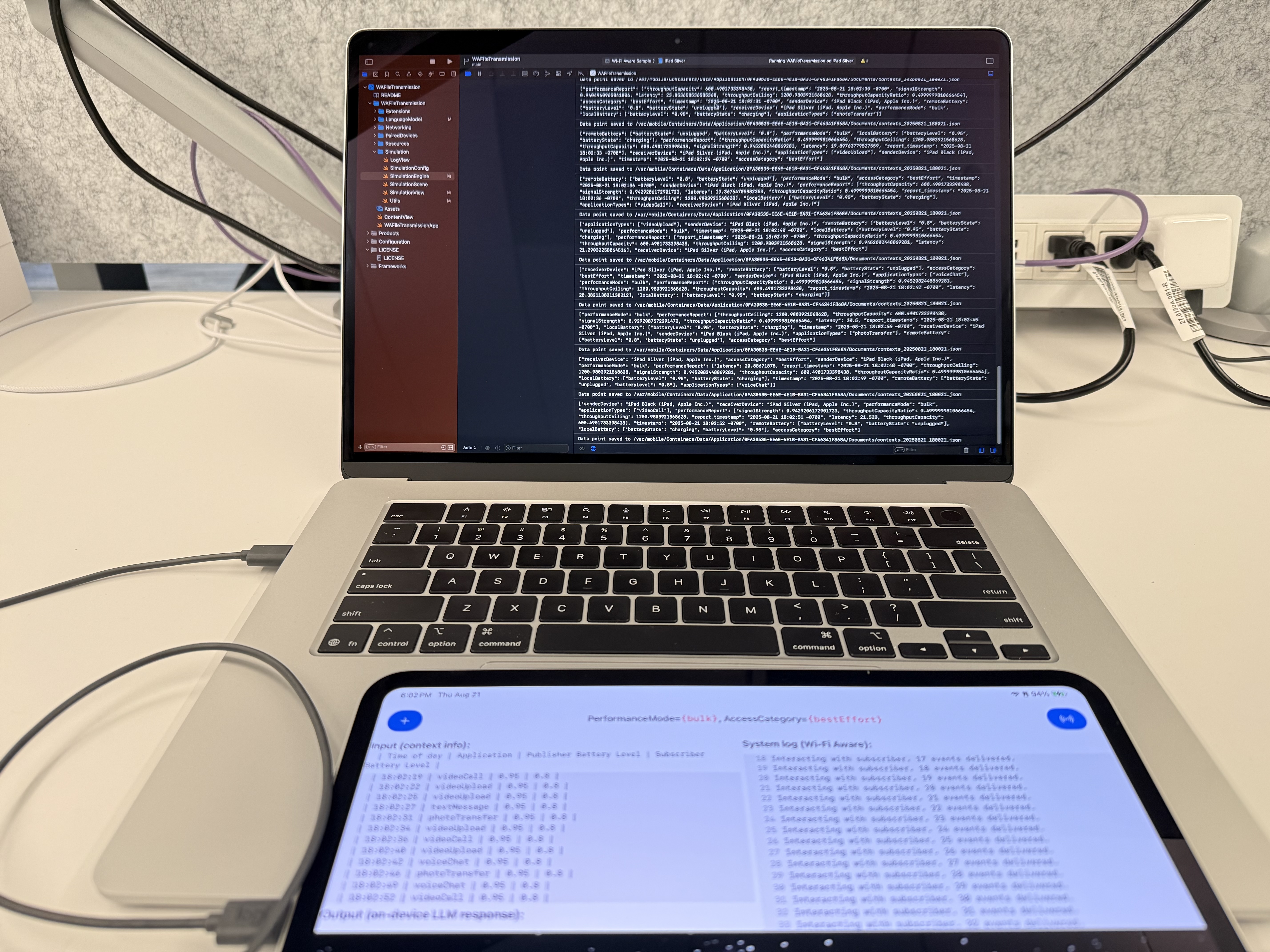}
\caption{Data collection pipeline. WA logs are recorded on the publisher and uploaded to a computer for pre-processing.}
\label{fig:data-realworld}
\end{figure}

\begin{figure}[!h]
\centering
\includegraphics[width=.55\textwidth]{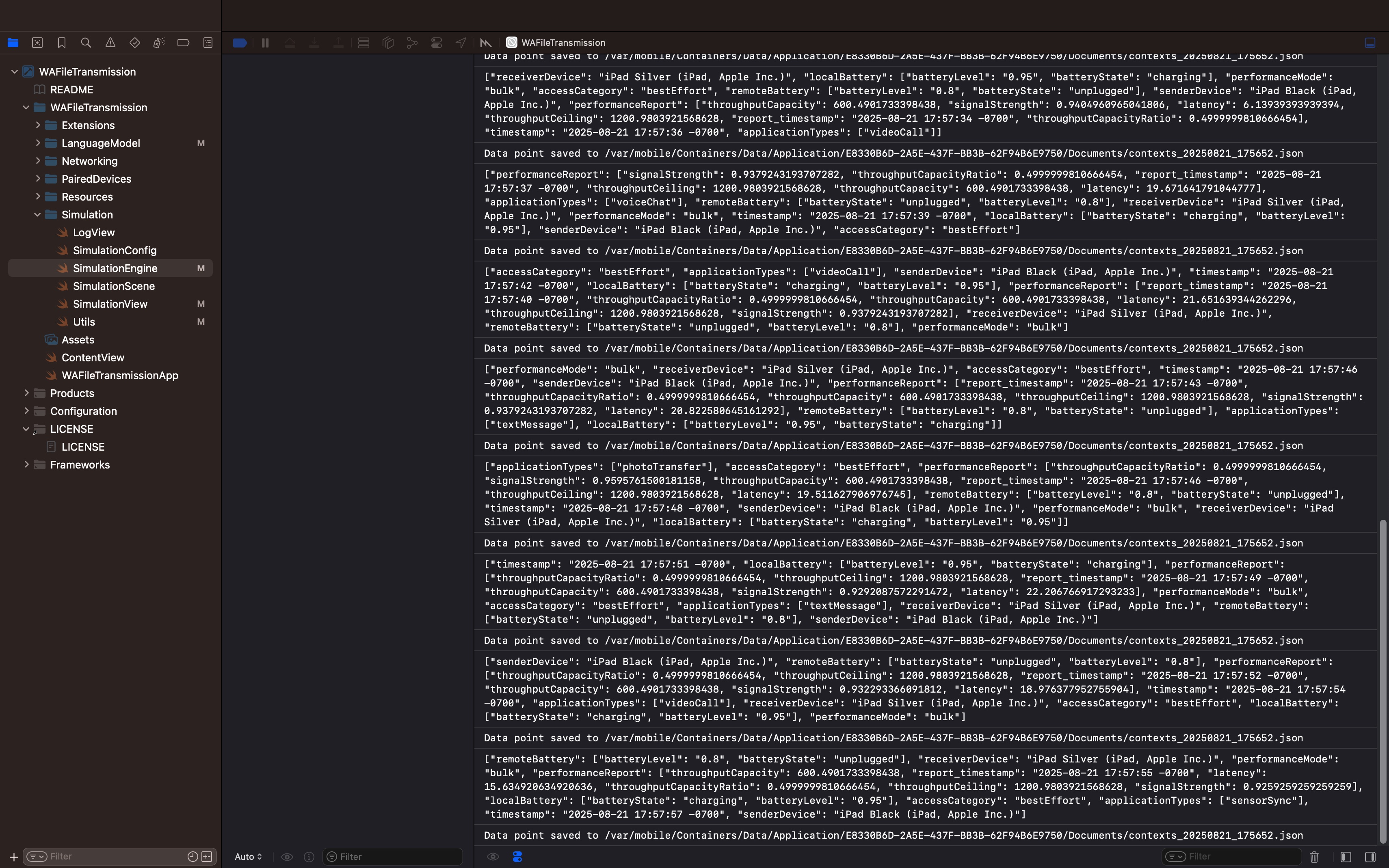}
\caption{Example WA logs received by the computer, including device context, WA parameters, and performance metrics.}
\label{fig:data-logs}
\end{figure}

\BfPara{Logged features}  
The following features were logged per record: 
\begin{itemize}
    \item Device context: publisher and subscriber device IDs, battery levels, time of day, application type.
    \item WA parameters: current $\texttt{PerformanceMode}$ and $\texttt{AccessCategory}$.
    \item Performance metrics: measured latency and battery usage.
\end{itemize}
Additional fields such as charging status, throughput capacity, signal strength, and throughput-related counters were also recorded (Figs.~\ref{fig:data-realworld}, \ref{fig:data-logs}), but were not used for training or inference. These remain available for future extensions or cross-validation.  

\newpage
\section{Baseline Details (\textbf{Rule})}\label{sec:baseline}

The \textbf{Rule} baseline is a simple heuristic policy that depends only on application type. 
For each application, we assign a preferred WA parameter tuple based on common-sense intuition about its latency or throughput requirements (Table~\ref{tab:preferred}).  
At inference time, the baseline examines the past $n$ applications (with $n=10$ in our setup, consistent across all baselines) and selects the parameter tuple that appears most frequently among their preferred mappings.  

This design makes \textbf{Rule} adaptive only to recent application history, without considering device battery states or other contextual factors. It thus serves as a lightweight non-learning baseline for comparison against context-aware methods such as PEARL and PEARL-Lite. Future extensions could incorporate a simple “battery-safe” heuristic—e.g., switching to energy-preserving modes below 20\% battery—to better reflect real-world smartphone behavior.


\begin{table}[!h]   
\centering
\resizebox{.5\columnwidth}{!}{
\centering
\begin{tabular}{lc}
\toprule
\textbf{Application Type} & \textbf{Preferred Parameter Tuple} \\
\midrule
\texttt{textMessage}     & (\texttt{realtime}, \texttt{bestEffort}) \\
\texttt{voiceChat}       & (\texttt{realtime}, \texttt{interactiveVoice}) \\
\texttt{videoCall}       & (\texttt{realtime}, \texttt{interactiveVideo}) \\
\texttt{sensorSync}      & (\texttt{bulk}, \texttt{background}) \\
\texttt{photoTransfer}   & (\texttt{bulk}, \texttt{bestEffort}) \\
\texttt{videoUpload}     & (\texttt{bulk}, \texttt{bestEffort}) \\
\texttt{firmwareUpdate}  & (\texttt{bulk}, \texttt{background}) \\
\texttt{mapSync}         & (\texttt{realtime}, \texttt{bestEffort}) \\
\bottomrule
\end{tabular}
}
\vspace{0.3em}
\caption{Preferred WA parameter tuple assigned heuristically for each application type.}
\label{tab:preferred}
\end{table}

\newpage
\section{Single-Objective Optimization Results}\label{sec:single-objective}

\begin{figure}[!h]
\centering
\begin{subfigure}[b]{0.3\textwidth}
    \includegraphics[width=\textwidth]{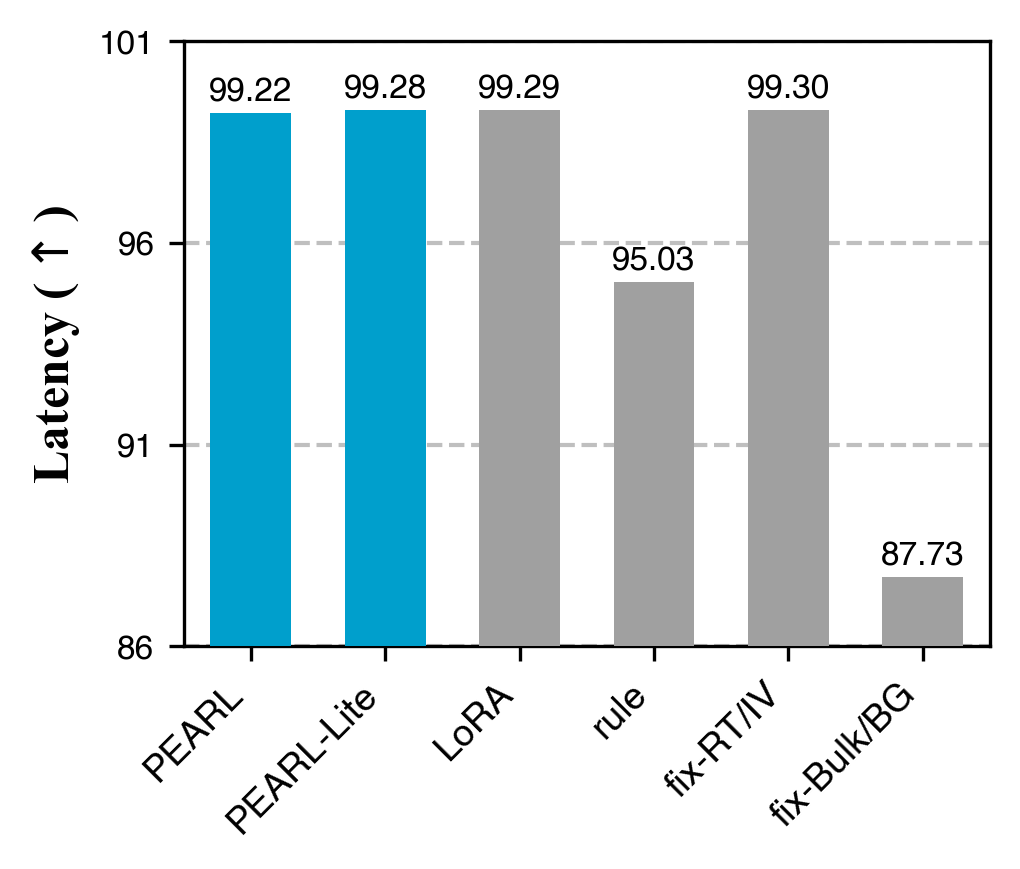}
    \caption{Latency-only}
    \label{fig:objective-latency}
\end{subfigure}
\hspace{0.5em}
\begin{subfigure}[b]{0.3\textwidth}
    \includegraphics[width=\textwidth]{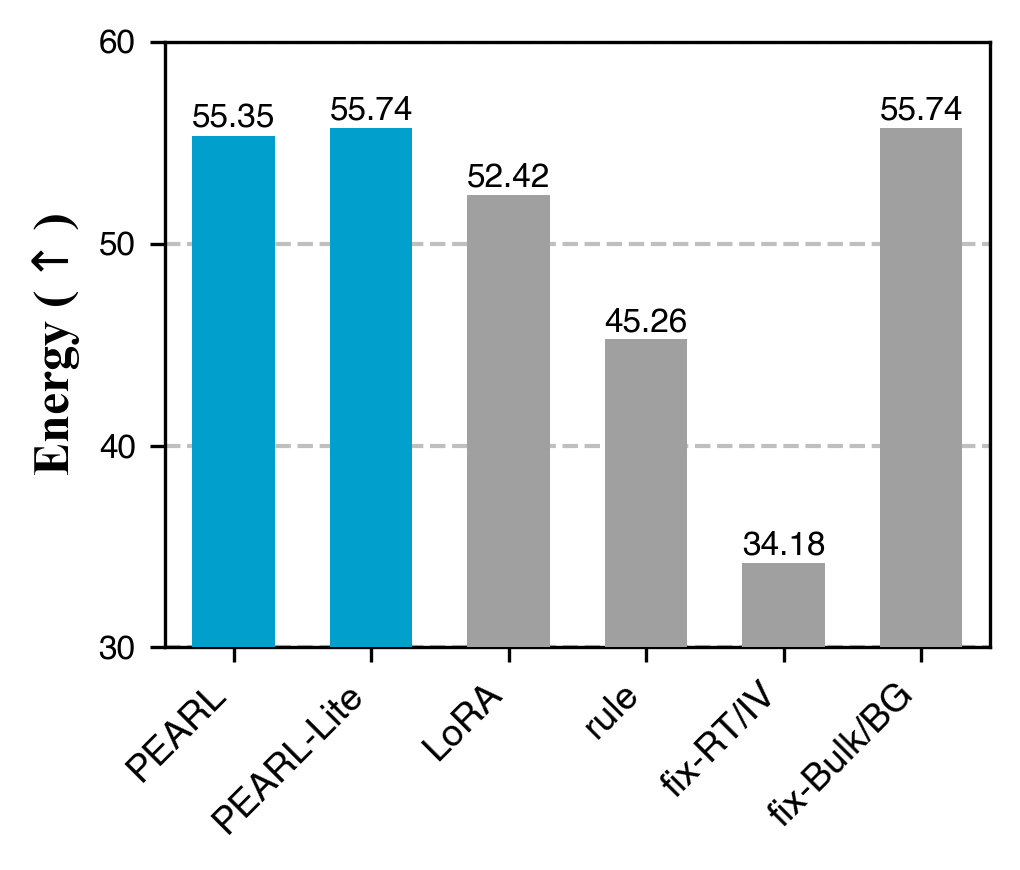}
    \caption{Energy-only}
    \label{fig:objective-energy}
\end{subfigure}
\caption{Performance comparison of PEARL variants and baselines for single-objective optimization.}
\label{fig:single-objective}
\end{figure}

Although PEARL is primarily designed for the joint optimization of latency and energy, we also evaluate performance under single-objective settings. 
Figure~\ref{fig:single-objective} compares PEARL variants against baselines when optimizing for \textbf{Latency} only or \textbf{Energy} only.  

In the latency-only setting (Fig.~\ref{fig:objective-latency}), \textbf{PEARL}, \textbf{PEARL-Lite}, and \textbf{LoRA-only} achieve scores within 1\% of \textbf{Fixed(RT/IV)}, which always selects $(\texttt{realtime}, \texttt{interactiveVoice})$—a configuration strongly favoring low latency.  
In the energy-only setting (Fig.~\ref{fig:objective-energy}), \textbf{PEARL} and \textbf{PEARL-Lite} achieve performance close to \textbf{Fixed(Bulk/BG)}, the parameter choice that minimizes energy consumption, while outperforming other adaptive baselines.  

Interestingly, \textbf{PEARL-Lite} performs on par with, and sometimes slightly better than, full \textbf{PEARL} in single-objective settings. 
This suggests that lightweight head-only models may be particularly well-suited for simpler optimization tasks, offering both efficiency and strong performance.

\newpage
\section{Head Module Design Details}\label{sec:ablation-head}

The head module in \textbf{PEARL-Lite} (and \textbf{PEARL}) is responsible for mapping the last-token embedding of the frozen LLM encoder to logits over the WA action space. 
Because this component is lightweight compared to the backbone LLM, its depth can be adjusted without incurring meaningful additional cost. 
We therefore explore different head architectures to examine whether deeper designs improve representational power.  

\begin{table}[!h]   
\centering
\resizebox{\columnwidth}{!}{
\centering
\begin{tabular}{l|ccc|cc|cc}
\toprule
\multirow{2}{*}{\textbf{Variant}} & \multicolumn{3}{c|}{\textbf{Performance} $\uparrow$} & \multicolumn{2}{c|}{\textbf{Inference} $\downarrow$} & \multicolumn{2}{c}{\textbf{Training} $\downarrow$} \\
\cmidrule(lr){2-4} \cmidrule(lr){5-6} \cmidrule(lr){7-8}
& \textbf{Objective Score} & \textbf{Latency} & \textbf{Energy} & \textbf{Time (ms)} & \textbf{RAM (GB)} & \textbf{Time (min/epoch)} & \textbf{RAM (GB)} \\
\midrule
3-layer Head & 7.54 & 99.27 & 41.85 & 14.5 & 6.80 & 12 & 11.03  \\
2-layer Head & 7.50 & 99.27 & 41.21 & 14.6 & 6.80 & 12 & 10.54 \\
1-layer Head & 7.51 & 99.27 & 41.35 & 14.7 & 6.80 & 12 & 11.30  \\
\bottomrule
\end{tabular}

}
\vspace{0.2em}
\caption{Performance and efficiency of different head module depths in \textbf{PEARL-Lite}.}
\label{tab:head-depth}
\end{table} 

As shown in Table~\ref{tab:head-depth}, the 3-layer head—the default configuration—achieves the highest objective score, while maintaining identical inference and training efficiency compared to shallower variants. 
This result suggests that slightly deeper heads can better utilize the LLM’s feature representations, while the overall system cost remains dominated by the backbone. 
Hence, deeper heads are a simple but effective design choice for improving performance without compromising efficiency.

\newpage
\section{Full Scenario Snapshots}\label{sec:snapshot}

To complement the quantitative analysis in Section~\ref{sec:impact_peer_info}, we present two qualitative snapshots of context logs and the corresponding predictions made by \textbf{PEARL}, \textbf{PEARL w/o peer info}, and \textbf{Rule}. 
These examples illustrate how peer-side context guides parameter selection toward more energy-aware configurations when the subscriber is low on battery.  

\begin{figure}[h]
    \centering
    \begin{subfigure}{\textwidth}
        \centering
        \includegraphics[width=0.9\textwidth]{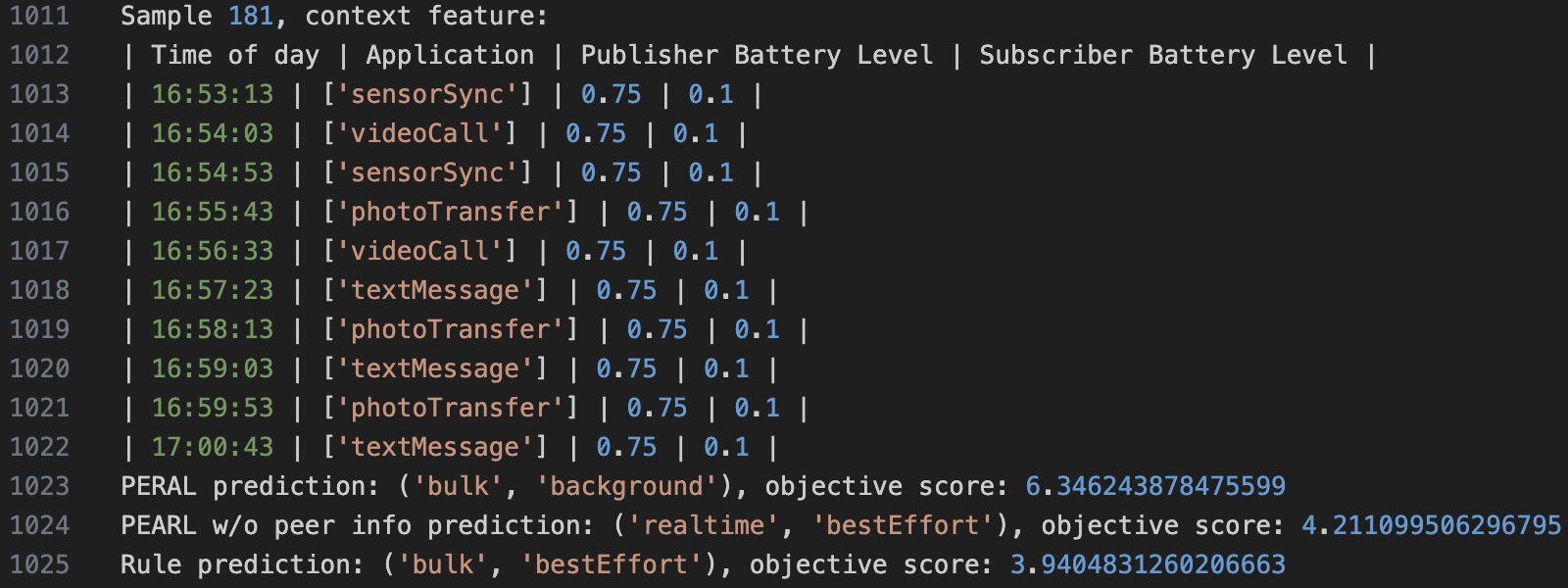}
        \caption{Afternoon scenario.}
        \label{fig:afternoon}
        \vspace{.5em}
    \end{subfigure}     
    \begin{subfigure}{\textwidth}
        \centering
        \includegraphics[width=0.9\textwidth]{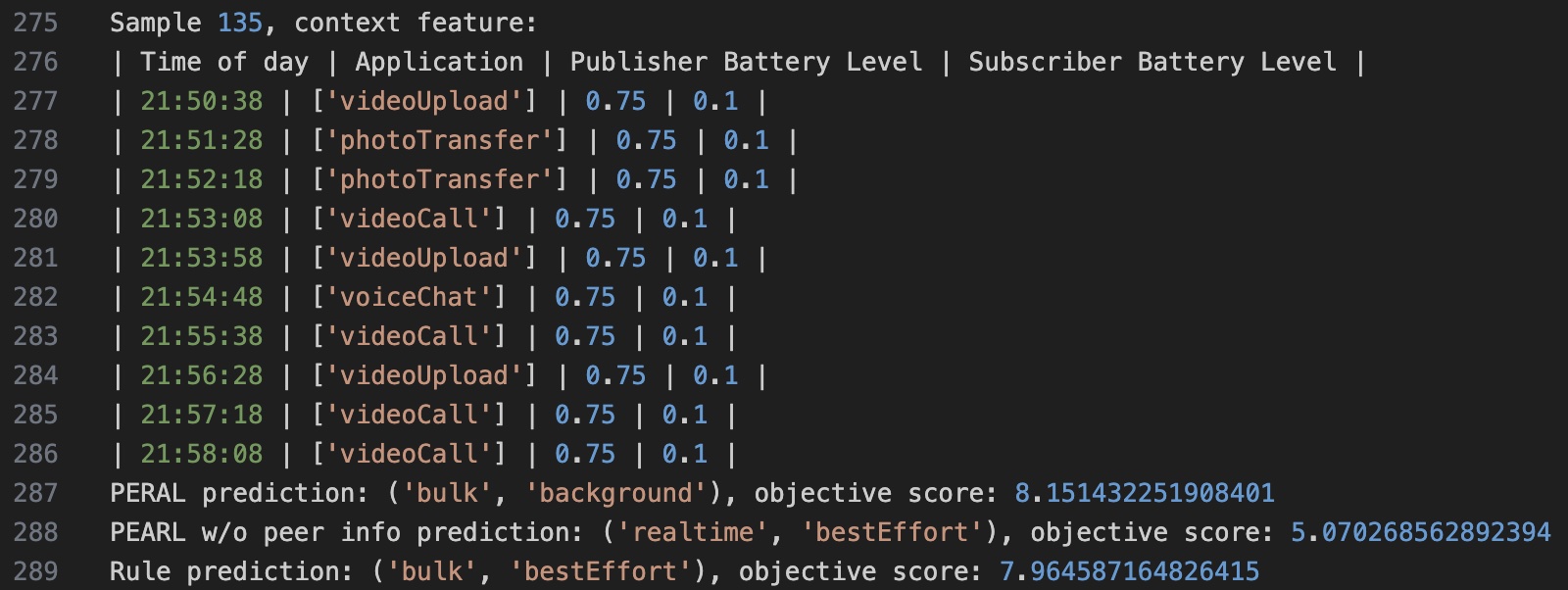}
        \caption{Evening scenario.}
        \label{fig:evening}
    \end{subfigure}
    \caption{Context logs and decisions of \textbf{PEARL}, \textbf{PEARL w/o peer info}, and \textbf{Rule} under the subscriber low-battery case.}
    \label{fig:snapshot}
\end{figure}

In Fig.~\ref{fig:afternoon}, when the subscriber battery is only 10\%, \textbf{PEARL} leverages this information to select \texttt{(bulk, background)}, prioritizing energy saving. 
Without access to peer context, \textbf{PEARL w/o peer info} observes only the publisher’s high battery level and instead selects \texttt{(realtime, bestEffort)}. 
\textbf{Rule}, based solely on application type, chooses \texttt{(bulk, bestEffort)}, yielding the lowest objective score.  

Fig.~\ref{fig:evening} shows a complementary case from the evening period. 
Here, despite a mix of latency-sensitive and high-throughput applications such as \texttt{videoCall} and \texttt{videoUpload}, \textbf{PEARL} again adapts to the low-battery subscriber and opts for \texttt{(bulk, background)}. 
Both \textbf{PEARL w/o peer info} and \textbf{Rule} fail to adjust accordingly, resulting in lower scores.  

Taken together, these snapshots demonstrate that \textbf{PEARL} consistently leverages peer-side context across different scenarios, leading to more energy-aware decisions while maintaining strong performance.








\end{document}